\documentclass[sigconf,authorversion,nonacm]{acmart}

\usepackage{times}
\usepackage{url}
\usepackage[utf8]{inputenc}
\usepackage{subcaption}
\usepackage{graphicx}
\usepackage{amsmath}
\usepackage{booktabs}
\usepackage{multirow}
\AtBeginDocument{%
  \providecommand\BibTeX{{%
    \normalfont B\kern-0.5em{\scshape i\kern-0.25em b}\kern-0.8em\TeX}}}

\setcopyright{none}
\copyrightyear{2022}
\acmYear{2022}

\begin{document}

\title{RHCO: A Relation-aware Heterogeneous Graph Neural Network with Contrastive Learning for Large-scale Graphs}

\author{Ziming Wan$^{1}$, Deqing Wang$^{1}$, Xuehua Ming$^{1}$, Fuzhen Zhuang$^{2}$, Chenguang Du$^{1}$, Ting Jiang$^{1}$, Zhengyang Zhao$^{1}$}
\affiliation{
    \institution{$^1$State Key Laboratory of Software Development Environment, School of Computer Science, Beihang University\country{China}. \\$^2$Institute of Artificial Intelligence, Beihang University\country{China}. }
}
\email{{buaa_wileloc, dqwang, xhming, zhuangfuzhen, duchenguang, royokong, zzy979}@buaa.edu.com}

\renewcommand{\shortauthors}{Ziming Wan, et al.}
\renewcommand{\shorttitle}{A Relation-aware Heterogeneous Graph Neural Network with Contrastive Learning for large-scale graphs}

\begin{abstract}
Heterogeneous graph neural networks (HGNNs) have been widely applied in heterogeneous information network tasks, while most HGNNs suffer from poor scalability or weak representation when they are applied to large-scale heterogeneous graphs. To address these problems, we propose a novel Relation-aware Heterogeneous Graph Neural Network with Contrastive Learning (RHCO) for large-scale heterogeneous graph representation learning. Unlike traditional heterogeneous graph neural networks, we adopt the contrastive learning mechanism to deal with the complex heterogeneity of large-scale heterogeneous graphs. We first learn relation-aware node embeddings under the network schema view. Then we propose a novel positive sample selection strategy to choose meaningful positive samples. After learning node embeddings under the positive sample graph view, we perform a cross-view contrastive learning to obtain the final node representations. Moreover, we adopt the label smoothing technique to boost the performance of RHCO. Extensive experiments on three large-scale academic heterogeneous graph datasets show that RHCO achieves best performance over the state-of-the-art models. Codes are available at \url{https://github.com/ZZy979/GNN-Recommendation.git}
\end{abstract}



\maketitle

\section{Introduction}

Heterogeneous graphs (HGs) are ubiquitous in real-world scenarios, such as social networks, bibliographic networks, academic networks, etc\cite{sun2013mining}. The characteristic that HGs contain various types of nodes and relations makes these graphs have abundant but hard-mining information. So learning node representations in HGs has emerged as a powerful strategy for analyzing graph data and could promote numerous tasks, including node classification\cite{metapath2vec, fu2017hin2vec}, ndoe clustering\cite{li2019spectral}, link prediction\cite{xue2020modeling, li2020type} and recommendation\cite{dong2012link}.

Recently, heterogeneous graph neural networks, which incorporate heterogeneity information into the message passing mechanism, have shown its superiority in handling heterogeneous graphs. For example, HGT \cite{hgt} uses type-specific parameters to model heterogeneity of HGs in the sampled sub-graph. HAN \cite{han} introduces the node-level attention and the semantic-level attention to discriminate significant neighbors in the metapath-based schema view. R-HGNN \cite{yu2022heterogeneous} explores the role of relations in the node representation learning, which also learns under the network schema view. However, most HGNNs perform their training procedure in a single view, which makes them suffer from insufficient usage of the information contained in HGs.

Contrastive learning, aimming to maximize the similarity between positive samples and minimize that between negative samples, is a widely used learning technique especially in the computer vision domain. It has the advantage that it can be performed across views. In this way, discriminative embeddings can be learned by fully using the information of heterogeneous graphs. However, few studies utilize contrastive learning in heterogeneous graph neural networks because the selection of proper views is a difficult problem. HeCo \cite{heco} is the first work to apply contrastive learning to heterogeneous graph representation learning, which adopts a cross-view contrastive learning between network schema view and metapath view to capture the local and higher-order structure information of heterogeneous graph. But HeCo cannot deal with large-scale graphs due to the unacceptable cost on constructing the metapath view.

Current heterogeneous graph neural networks have the following problems when applying to large-scale HGs: (1) \textbf{Poor scalability}. Most heterogeneous neural networks adopt pre-defined metapath to make the model learn meaningful semantic information. However, the number of metapath-based neighbors grows exponentially with the number of nodes in the graph, so these models cannot be applied to large-scale heterogeneous graphs \cite{yu2022heterogeneous}. (2) \textbf{Poor representation}. Large-scale heterogeneous graphs have complex structure and heterogeneity, which causes existing heterogeneous graph neural networks cannot learn efficient node representation in a single view of heterogeneous graphs.

To address the above problems, we propose a Relation-aware Heterogeneous Graph Neural Network with Contrastive Learning (RHCO) for large-scale heterogeneous graph representation learning.
We choose the network schema view and the positive sample graph view to perform a cross-view contrastive learning. For the network schema view, we design a graph encoder to learn relation-aware node embeddings. 
For the positive sample graph view, we propose a novel positive sample selection strategy to construct reasonable positive sample graphs, which also avoids the scalability problem, and then learn node embeddings of positive sample graph view. 
We obtain the final node representations by applying the contrastive learning across these two views. 
The network schema view can learn strong node representations by considering the role of different relations and the positive sample graph view can guide the learning process of the network schema view by providing the information of which pairs should be close and which should not.
Our main contributions are summarized as follows:
\begin{itemize}
\item We propose a relation-aware heterogeneous neural network with contrastive learning for large-scale graphs which captures the multiple relation-specific representations of heterogeneous graphs and employs contrastive learning mechanism to improve the learnt node representations. To our best knowledge, our approach is the first attempt to apply contrastive learning on large-scale heterogeneous graphs. 
\item We propose a positive sample selection strategy based on attention weights learned by a pre-trained model, which not only avoids explicit construction of metapath-based neighbor graph, but also improves the scalability of large-scale heterogeneous graph neural networks significantly.
\item We conduct extensive experiments on three large-scale public datasets and the results show our method outperforms SOTA models, which demonstrates the efficiency of RHCO on large-scale heterogeneous graphs.
\end{itemize}

\section{Related Work}

In this section, we review some existing literature related to our work, including heterogeneous graph learning, relation graph learning, and contrastive learning. Then we point out the differences between previous studies and our research.

\textbf{Heterogeneous Graph Learning}. Graph neural networks (GNNs) have attracted lots of interest, while most GNNs focus on homogeneous graphs \cite{wu2020comprehensive}. The basic idea of GNNs is first to propagate information among nodes and their neighbors, and then aggregate the received information to obtain node representations. Recently, some researchers have studied heterogeneous graphs and proposed several heterogeneous graph neural networks (HGNNs). For example, HAN \cite{han} converts a heterogeneous graph to multiple metapath-based neighbor graphs by predefining dataset-specific metapaths, then learns node embeddings by node-level attention and semantic-level attention. MAGNN \cite{fu2020magnn} improves HAN by considering intermediate nodes of metapath. 
Generally, these methods perform the message propagation and aggregation on metapath-based neighbors, so the learnt node embeddings contain semantic information. However, the above methods cannot be applied to large-scale datasets because of the unbearable cost of building meatapath-based neighbor graphs.

\textbf{Relation Graph Learning}. Relation graph learning methods study the role of relations. R-GCN \cite{rgcn} performs message passing by GCN on the relational bipartite graphs associated with target nodes to obtain node embeddings. RSHN \cite{zhu2019relation} construct edge-centric coarsened line graph to handle various relations and learn relation structure-aware node representations. RHINE \cite{Lu_Shi_Hu_Liu_2019} designs different distance functions for different relations to enhance the expressive power of node embeddings. HGT \cite{hgt} proposes heterogeneous mutual attention and message passing inspired by Transformer\cite{transformer} to perform on each relational bipartite. HGConv \cite{yu2020hybrid} uses GAT instead of GCN as micro-level convolution on each relational bipartite graph and learns the importance of relations by macro-level convolution. R-HGNN \cite{yu2022heterogeneous} improves HGConv by considering relational information in message passing, thus learning relation-aware node representations. However, the above methods failed to exploit supervised signals from the data itself to learn general node representations and suffer from the insufficient usage of the information in large-scale datasets.

\textbf{Contrastive Learning}. We mainly focus on reviewing graph-related contrastive learning. Specifically, DGI \cite{velickovic2019deep} uses attribute shuffle to generate the negative graph and utilizes Infomax to contrast the positive nodes, negative nodes, and global summaries. HDGI \cite{ren2020heterogeneous} improves DGI by using semantic-level attention to learn local node representations. Along this line, GMI \cite{peng2020graph} performs contrastive learning between the center node and its local path from node features and topological structure. DMGI conducts contrastive learning between original and corrupted networks on every metapath view and designs a consensus regularization to guide the fusion of different metapaths. HeCo \cite{heco} learns node embeddings from network schema view (relation-based) and metapath view (metapath-based) and performs contrastive learning across these two views. However, some existing methods perform contrastive learning in a single view, so they cannot capture the high-order factors in heterogeneous graphs. In contrast, other methods use metapaths to guide the construction of positive samples, which will consume too many resources in large-scale graphs.

Different from the methods above, we design a cross-view contrastive learning model which considers both local and high-level information of heterogeneous graphs and adopts the contrastive learning mechanism to supervise the node representation learning process. Remarkably, our approach could learn relation-aware node representations on large-scale heterogeneous graphs.

\begin{figure}
    \centering
    \includegraphics[width=1\columnwidth]{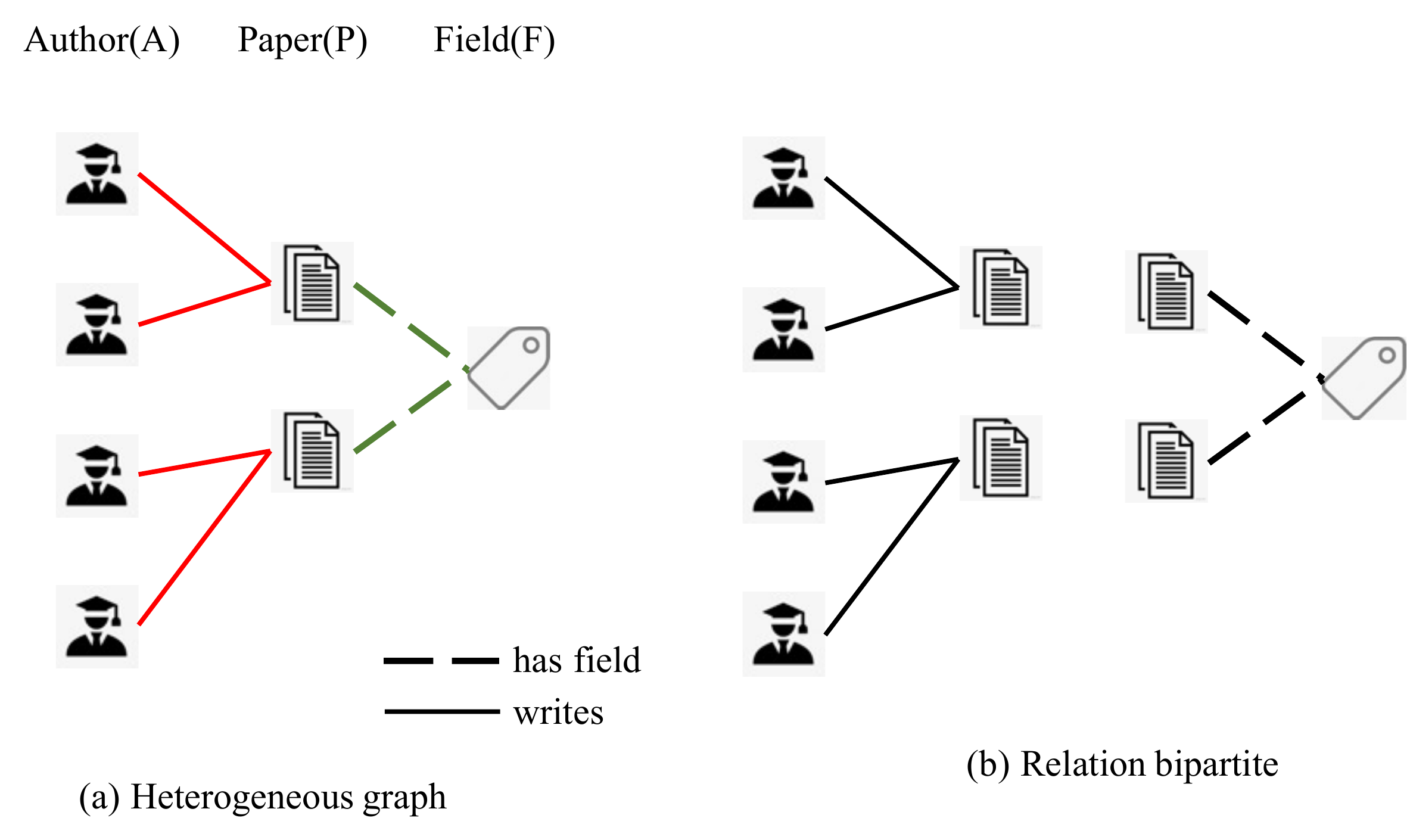}
    \vspace{-0.2cm}
    \caption{The illustration of heterogeneous graph, metapath and relation bipartite graph}
    \label{fig:define}
\end{figure}

\section{Preliminaries}
In this section, we formally define some significant concepts related to heterogeneous graphs as well as the formalization of the studied problem.

\textbf{Heterogeneous graph}. Heterogeneous graph is defined as a directed graph $G = (V, E)$, with node type mapping $\varphi: V\to A$ and edge type mapping $\psi: E \to R$, where $A$ and $R$ represent the set of node type and edge type respectively, and $|A| + |R|> 2$.

Figure \ref{fig:define} (a) illustrates a heterogeneous graph with three types of nodes, including author(A), paper(P) and field(F). There are two types of edges, which are author \textit{writes} paper and paper \textit{has field}.

\textbf{Relation}. For an edge $e = (u, v)$ linked from source node $u$ to target node $v$, the corresponding relation is $r = <\varphi(u), \psi(e), \varphi(v)>$.

\textbf{Relational bipartite graph}. Given a heterogeneous graph $G$ and a relation $r$, the bipartite graph $G_r$ is defined as a graph composed of all the edges of the corresponding type of the relation $r$.

Figure \ref{fig:define} (b) are the relational bipartite graphs of Figure \ref{fig:define} (a). Specifically, the relational bipartite graph of relation \textit{writes} contains two types of nodes: authors and papers. The relational bipartite graph of relation \textit{has field} contains papers and fields accordingly.

\textbf{Metapath}. Metapath $P$ is defined as a path with the following form: $A_1 \xrightarrow{R_1} A_2 \xrightarrow{R_2} \cdots \xrightarrow{R_{l-1}} A_l$ (abbreviated as $A_1 A_2 \cdots A_l$), where $A_i \in A, R_i \in R$. The metapath describes a composite relation between node types $A_1$ and $A_l$, which expresses specific semantics.

Figure \ref{fig:define} (a) contains many metapaths including PAP and PFP. PAP describes that two papers are written by the same author, which is highlighted by red color in Figure \ref{fig:define} (a). PFP describes that two papers have the same field, shown by the green color.

\textbf{Metapath-based neighbor graph}. Given a metapath $P$ of a heterogeneous graph $G$, the neighbor graph $G_P$ is defined as a graph composed of all neighbor pairs based on metapath $P$. If $P$ is symmetric, $G_P$ is a homogeneous graph, otherwise $G_P$ is a bipartite graph containing two types of nodes $A_1$ and $A_l$.

\textbf{Heterogeneous graph representation learning}. Given a graph $G = (V, E)$, graph representation learning aims to learn a function $f: V \to R^d, d \ll |V|$ to map the nodes in the graph to a low-dimensional vector space while preserving the topological structure information of the graph. These embedding vectors can be used for a variety of downstream tasks, such as node classification, node clustering and link prediction.

\begin{figure*}
    \centering
    \includegraphics[width=2\columnwidth]{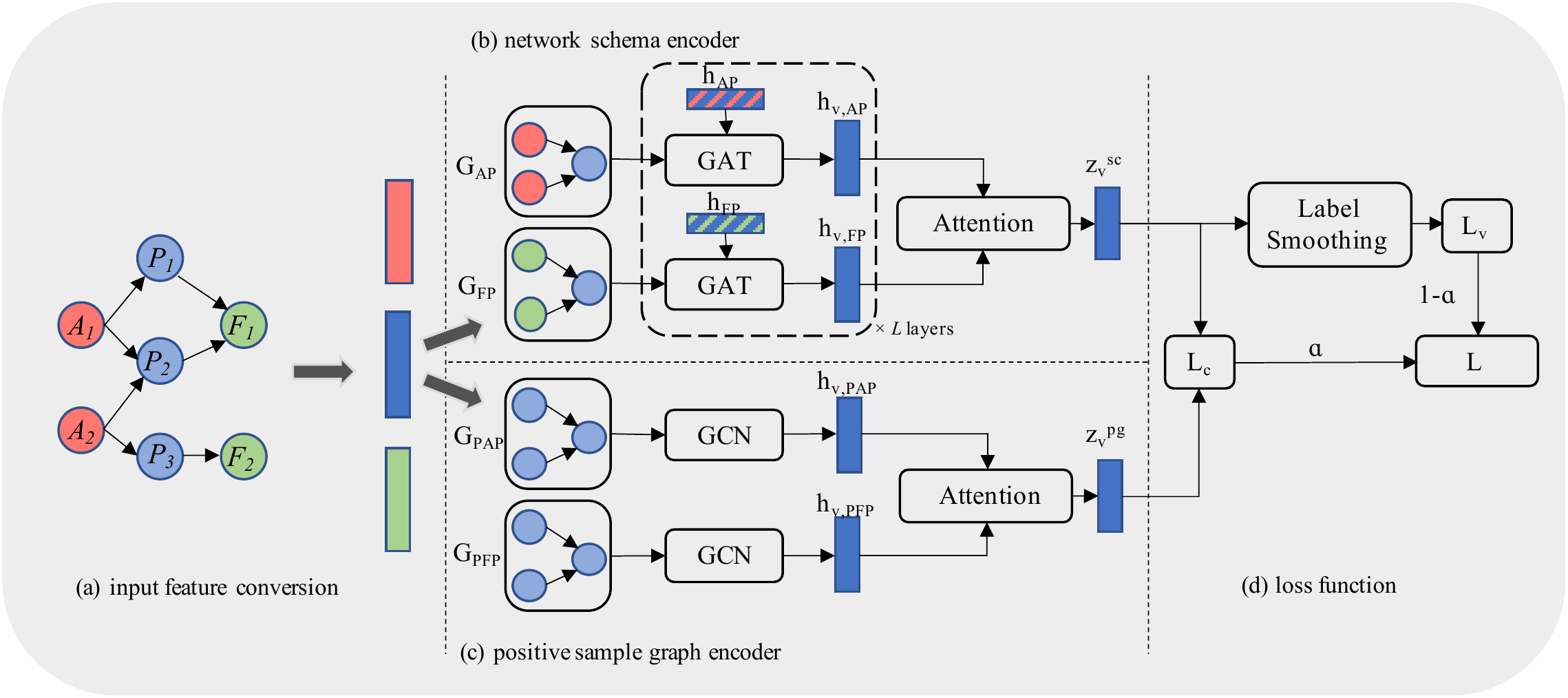}
    \vspace{-0.2cm}
    \caption{Framework of RHCO model. RHCO respectively learns node embeddings $z_v^{sc}$ and $z_v^{pg}$ from the network schema encoder and positive sample graph encoder then performs contrastive learning to obtain final node representations.}
    \label{fig:RHCO}
\end{figure*}

\section{Methodology}

In this section, we propose our model RHCO, which takes a sampled graph $G$ as input and outputs low-dimensional node representations of target nodes. The framework of RHCO is shown in Figure \ref{fig:RHCO}. RHCO consists of four components: input feature conversion, network schema encoder, positive sample graph encoder and contrastive loss function.
Network schema encoder and positive sample graph encoder capture both local structure and high-order semantic information of the graph respectively, thus making full use of hidden information of the heterogeneous graph. The outputs of the two encoders are used to calculate contrastive loss and the output of network schema encoder is processed by C\&S label smoothing \cite{cands} and then used as the input of the downstream classifier to obtain the classification loss. Train procedure is guided by the weighted sum of contrastive loss and classification loss.

For a given heterogeneous bibliographic network $G$ consisting of three types of nodes: author (A), paper (P) and field (F), we convert $G$ to relational bipartite graphs ($G_{AP}$, $G_{FP}$) and metapath-based neighbor graphs ($G_{PAP}$, $G_{PFP}$) to learn node embeddings respectively, and then perform contrastive learning across these two views to supervise each other. 
We will introduce the components of RHCO step by step.



\subsection{Input Feature Conversion}

We project the input features of all types of nodes into the same latent vector space with type-specific linear transformations to eliminate the bias of different types of nodes, as shown in Figure \ref{fig:RHCO} (a). And inspired by R-HGNN \cite{yu2022heterogeneous}, we consider the relation information in the graph representation learning.
For node $v$ and its associated relation $r$, their representations are converted as follows:
\begin{equation}
\label{eq:input_feature_conversion}
\begin{aligned}
& z_{v,r}^l = W_{\varphi(v)}^l h_{v,r}^{l-1} \\
& z_r^l = W_r^l h_r^{l-1}
\end{aligned}
\end{equation}
where $h_{v,r}^l \in R^d$ and $h_r^l \in R^{d_{rel}}$ are the $l$-th level representation of node $v$ under relation $r$ and the $l$-th level representation of relation $r$ itself, $h_{v,r}^0$ is set as the input feature $x_v$ of node $v$, $h_r^0$ is one-hot encoding, $W_{\varphi(v)}^l$ and $W_r^l$ are type-specific transformation matrices.

\subsection{Network Schema Encoder}

Network schema encoder aims to learn the embeddings of target nodes under the network schema view, and capture the local structure of the graph by aggregating information from different types of first-order neighbors, as shown in Figure \ref{fig:RHCO} (b).

Since different types of neighbors, as well as different neighbors of the same type, have different contributions to the target node, we use the attention mechanism to aggregate the information of neighbors of the same type and different types respectively.

First, we use node level attention to aggregate information from neighbors of relation $r$ at the $l$-th layer:
\begin{equation}
\label{eq:node_level_attention}
\begin{aligned}
& \alpha_{u,v}^{r,l} = \frac{\exp(LeakyReLU((z_r^l)^T [z_{u,r}^l || z_{v,r}^l]))}{\sum_{u' \in N_v^r} \exp(LeakyReLU((z_r^l)^T [z_{u',r}^l || z_{v,r}^l]))} \\
& h_{v,r}^l = \sigma(\sum_{u \in N_v^r} \alpha_{u,v}^{r,l} z_{u,r}^l)
\end{aligned}
\end{equation}
where $R_v$ is the set of relations associated with target node $v$, $N_v^r$ is the set of neighbors of node $v$ under relation $r$, $\alpha_{u,v}^{r,l}$ is the importance of neighbor $u$ under relation $r$ to target node $v$, $||$ represents vector concatenation, and $\sigma$ is activation function. We use representation vectors of relation instead of the learnable attention vector to capture information of relations, so the embeddings are relation-aware.

To enhance the learning process, we employ the multi-head attention mechanism, which repeats Eq.~\ref{eq:node_level_attention} $K$ times and concatenates the $K$ output vectors of the attention heads.
Eqs. \ref{eq:input_feature_conversion} and \ref{eq:node_level_attention} describe the message passing process of a single layer. The embedding of target node $v$ under all relations $\{h_{v,r}^L | r\in R_v\}$ are obtained by stacking $L$ layers. Then we use type-level attention to combine these embeddings to get the output of network schema encoder:
\begin{equation}
\label{eq:type_level_attention}
\begin{aligned}
& \beta_{v,r} = \frac{\exp(LeakyReLU((U_r h_{v,r}^L)^T (V_r h_r^L)^T))}{\sum_{r' \in R_v} \exp(LeakyReLU((U_{r'} h_{v,r'}^L)^T (V_{r'} h_{r'}^L)^T))} \\
& z_v^{sc} = \sum_{r \in R_v} \beta_{v,r} U_r h_{v,r}^L
\end{aligned}
\end{equation}
where $\beta_{v,r}$ is the importance of the relation $r$ to target node $v$, $U_r$ and $V_r$ are transformation matrices of the node representation and the relation representation, respectively.

Since Eq. \ref{eq:type_level_attention} only considers the first-order neighbor relation set $R_v$, the embeddings of target node $v$ under the network schema view captures local structure information of the heterogeneous graph.

\begin{figure}
    \centering
    \includegraphics[width=1\columnwidth]{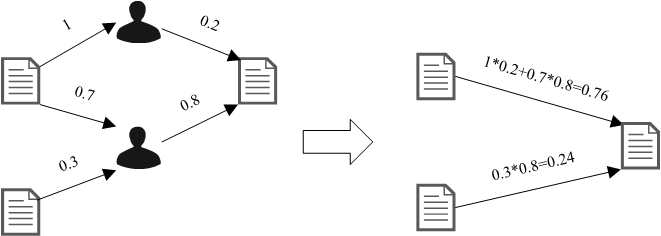}
    \caption{The illustration of the proposed positive sample selection strategy}
    \label{fig:positive_sample_graph}
    \vspace{-0.5cm}
\end{figure}

\subsection{Positive Sample Selection Strategy}
Before introducing the positive sample graph encoder, we first discuss how we select positive samples of a target node, as well as avoid the scalability problem troubling the existing methods. The main idea is to select essential nodes as positive samples in linear time. So we use attention weight calculated by a pre-trained model, which can reflect the critical nodes of a target node, to select a fixed number of nodes of the same node type as positive samples for each target node. Also, the attention weights are set once finished the pre-train so we can get the positive samples in linear time.

For target node $v$ and node $u$ of the same type connected by intermediate node $u'$ of type $\varphi$, the attention weight of $u$ to $v$ is calculated as follows:
\begin{equation}
\label{eq:positive_sample_weight}
a_{u,v}^\varphi = \sum_{u' \in N_v^\varphi} e_{u,u'} e_{u',v}
\end{equation}
where $e_{u,v}$ is the attention weight of node $u$ to node $v$ calculated by the pre-trained model and $N_v^\varphi$ is the set of neighbors of node $v$ with type $\varphi$. For the target node $v$, $T_{pos}$ nodes of the same type with the largest attention weight are selected as its positive samples, as shown in Figure \ref{fig:positive_sample_graph}. Positive sample graph is formed by all pairs of positive samples.

Positive sample graph is a homogeneous graph consisting of target nodes, and its edges contain metapath information. Because the intermediate node $u'$ in Eq. \ref{eq:positive_sample_weight} is the first-order neighbor of type $\varphi$ of the target node $v$, and node $u$ is the second-order neighbor of $v$, node $u$ is essentially a neighbor of node $v$ based on metapath $p = \varphi_v \varphi \varphi_v$. Therefore, positive sample graph is essentially a metapath-based neighbor graph. For example, if the target node in Figure \ref{fig:positive_sample_graph} is paper (P), the paper nodes connected to the target node through scholar (A) node are neighbors based on metapath $PAP$.

For each neighbor types $\Phi_v$ associated with the target node, a positive sample graph is constructed separately: $\{G_{\varphi_v \varphi \varphi_v} | \varphi \in \Phi_v\}$, used as input graphs of positive sample graph encoder.

In addition, we calculate an overall weight of neighbor $u$ by summing all attention weights of $u$ to 
$v$ connected by type $\varphi$ and select $T_{pos}$ nodes with the largest weight to construct an overall positive sample graph $G_{pos}$:
\begin{equation}
a_{u,v} = \sum_{\varphi \in \Phi_v} a_{u,v}^\varphi
\end{equation}
This overall positive sample graph comprehensively considers all metapaths, and is used for calculating contrastive loss and label smoothing.

Compared with the metapath-based positive sample selection strategy adopted by HeCo \cite{heco}, the proposed positive sample selection strategy does not explicitly construct metapath-based neighbor graphs to get the number of metapaths connecting two nodes, but calculate positive samples for each target node using pre-trained attention weights, which only considers the first and second order neighbors. So the proposed strategy can not only be used on large-scale graphs but also select better positive samples.

\subsection{Positive Sample Graph Encoder}

Positive sample graph encoder aims to learn the embeddings of target nodes under positive sample view, and capture high-order structure information of the graph by aggregating information from neighbors in positive sample graphs, as shown in Figure \ref{fig:RHCO} (c). After obtaining positive sample graphs, we perform message passing on these graphs to learn the node embeddings under positive sample view.

Suppose the set of neighbor types associated with target node $v$ is $\Phi_v$, and the corresponding metapath set is $P_v = \{\varphi_v \varphi \varphi_v | \varphi \in \Phi_v\}$ that end at node $v$. 
The metapath-based neighbors of node $v$ based on the metapath $p = \varphi_v \varphi \varphi_v$ is the set of neighbors in the corresponding positive sample graph $G_p$, and it is also the set of positive samples of node $v$. 

For metapath $p$, we use single-layer GCN on the corresponding positive sample graph $G_p$ to aggregate information from positive samples:
\begin{equation}
\label{eq:pg_encoder_gcn}
h_{v,p} = \frac{1}{|N_v^p|} \sum_{u \in N_v^p} W_p x_u
\end{equation}
where $x_u$ is the input feature of node $u$, and $W_p$ is metapath-specific transformation matrix.

After obtaining the embedding of target node $v$ under all metapaths $\{h_{v,p} | p \in P_v\}$, we use semantic level attention to combine these embeddings to obtain the output of the positive sample graph encoder $z_{pg}$:
\begin{equation}
\begin{aligned}
& w_p = \frac{1}{|V|} \sum_{v \in V} q^T \cdot \tanh(W^{pg}h_{v,p} + b^{pg}) \\
& \beta_p = \frac{\exp(w_p)}{\sum_{p' \in P_v} \exp(w_{p'})} \\
& z_v^{pg} = \sum_{p \in P_v} \beta_p h_{v,p}
\end{aligned}
\end{equation}
where $\beta_p$ is the importance of metapath $p$ to target node $v$, $W^{pg}$ and $b^{pg}$ are learnable parameters, $q$ is the semantic level attention vector.

As mentioned above, the positive sample pairs are essentially metapath-based neighbors, so the embeddings of target node $v$ under positive sample view capture the high-order structure information of the graph.





\subsection{Loss Function}

After obtaining the node embeddings $z_v^{sc}$ and $z_v^{pg}$ under the above two views, we use a two-layer fully connected network to perform linearly transform, and then calculate contrastive loss $\mathcal{L}_c$ as follows:
\begin{equation}
\label{eq:contrastive_loss}
\begin{aligned}
& \mathcal{L}_v^{sc} = -\log \frac{\sum_{u \in N_v} \exp(\cos(z_v^{sc},z_u^{pg}) / \tau)}{\sum_{u' \in V} \exp(\cos(z_v^{sc},z_{u'}^{pg}) / \tau)} \\
& \mathcal{L}_v^{pg} = -\log \frac{\sum_{u \in N_v} \exp(\cos(z_v^{pg},z_u^{sc}) / \tau)}{\sum_{u' \in V} \exp(\cos(z_v^{pg},z_{u'}^{sc}) / \tau)} \\
& \mathcal{L}_c = \frac{1}{|V|} \sum_{v \in V} (\lambda \mathcal{L}_v^{sc} + (1 - \lambda) \mathcal{L}_v^{pg})
\end{aligned}
\end{equation}
where $N_v$ is the set of neighbors of the target node $v$ in positive sample graph $G_{pos}$, the rest nodes are regarded as negative samples of node $v$, $\tau$ is temperature parameter of the contrastive loss, and $\lambda$ is balance coefficient. 
When using mini-batch training, $V$ in Eq. \ref{eq:contrastive_loss} denotes the set of all target nodes in a batch. Note that it requires not only the embeddings of target nodes, but also that of their positive samples to calculate contrastive loss. Therefore, when calculating the message passing of each batch, the actual target nodes should be $V \cup \{u \in N_v | v \in V\}$.

For the node classification task, we use node labels to guide the end-to-end training procedure. We follow the basic idea of label propagation proposed in C\&S \cite{cands} and perform label smoothing on positive sample graph $G_{pos}$. 
We use an MLP layer, whose output dimension is the number of classes, to obtain the predict labels of test set from the output of network schema encoder $z_{sc}$. Along with real labels of training set, we construct basic prediction $G \in R^{|V| \times d}$. Then we perform multiple iterations:

\begin{equation}
\begin{aligned}
& G^{(0)} = G \\
& G^{(t + 1)} = \gamma SG^{(t)} + (1 - \gamma)G
\end{aligned}
\end{equation}
where $G^{(t)}$ is smoothed prediction at iteration $t$, $S = D^{-1/2}AD^{-1/2}$ is the normalized adjacency matrix of the label propagation graph, $\gamma$ is weight coefficient.
After $T$ iterations, the final prediction $\hat{Y} = G^{T} $ is used as input of the downstream classifier.
The classification loss is calculated by cross-entropy:
\begin{equation}
\mathcal{L}_v = -\sum_{v \in V} \sum_{c=1}^C y_{v,c} \log \hat{y}_{v,c}
\end{equation}
The final loss is the combination of contrastive loss $\mathcal{L}_c$ and classification loss $\mathcal{L}_v$ by weight $\alpha$ :
\begin{equation}
\mathcal{L} = \alpha \mathcal{L}_c + (1 - \alpha) \mathcal{L}_v
\end{equation}

\subsection{Analysis of Model Complexity}

The proposed RHCO is efficient on large-scale heterogeneous graphs. Suppose the input and output dimension of node embeddings at the $l$-th layer of the network schema encoder are $N^l_{in}$ and $N^l_{out}$. Let $R^l_{in}$ and $R^l_{out}$ denote the input and output dimension of relation representations at the $l$-th layer. The time complexity of aggregating information from neightbors of relation $r \in R$ is linear to the number of nodes and edges in the corresponding relational bipartite graph $G_r$. It can be represented by $O(\alpha|V_r|+\beta|E_r|+\gamma)$, where $|V_r|$ and $|E_r|$ are the number of nodes and edges in the $G_r$. $\alpha = N^l_{out}(N^l_{in}+|R|)$, $\beta = R^l_{in}N^l_{out}$ and $\gamma = R^l_{in}R^l_{out}$. When combining the relation-specific node embeddings, the time complexity is linear to the number of nodes in the heterogeneous graph $G$, which can be denoted as $O(d|R||V|(R^L_{out}+N^L_{out}))$, where $d$ is the demension of final compact node representations and $L$ is the number of layers. The time complexity of calculating node representations under the positive sample graph view is linear to the number of nodes in the positive sample graph $G_{p}$. It can be indicated by $O(N_{in}N_{out}|V_p|)$, where $N_{in}$ and $N_{out}$ are the input and output dimension of positive sample graph encoder.

\section{Experiments}

\subsection{Dataset}

We employ three real-world, large-scale heterogeneous academic network datasets for node classification. The basic information is summarized in Table \ref{tab:dataset}.


\begin{table}
  \newcommand{\tabincell}[2]{\begin{tabular}{@{}#1@{}}#2\end{tabular}}
  \centering
  \caption{Statistics of Datasets}
  \label{tab:dataset}
    \begin{tabular}{|c|c|c|}
      \hline
      Dataset & \#Nodes & \#Edges \\
      \hline
      \multirow{4}{*}{ogbn-mag} & Author(A): 1,134,649 & P-A: 7,145,660 \\
      & Paper(P): 736,389 & P-P: 5,416,271 \\
      & Field(F): 59,965 & P-F: 7,505,078 \\
      & Institution(I): 8,740 & A-I: 1,043,998 \\
      \hline
      \multirow{4}{*}{oag-venue} & Author(A): 2,248,205 & P-A: 6,349,317 \\
      & Paper(P): 1,852,225 & P-P: 9,194,781 \\
      & Field(F): 120,992 & P-F: 17,250,107 \\
      & Institution(I): 13,747 & A-I: 1,726,212 \\
      \hline
      \multirow{4}{*}{oag-field} & Author(A): 2,248,205 & P-A: 2,619,759 \\
      & Paper(P): 714,192 & P-P: 1,718,037 \\
      & Venue(V): 11,177 & P-V: 714,192 \\
      & Institution(I): 13,747 & A-I: 1,726,212 \\
      \hline
    \end{tabular}
\end{table}


\begin{table*}
  \newcommand{\tabincell}[2]{\begin{tabular}{@{}#1@{}}#2\end{tabular}}
  \centering
  \caption{Results on Node Classification}
\label{tab:result}
    \begin{tabular}{|c|c|c c c c c c c c|}
      \hline
      Dataset & Metric & GAT & R-GCN & C\&S & Mp2vec & HGT & HGConv & R-HGNN & RHCO \\
      \hline
      \multirow{2}{*}{ogbn-mag} & Accuracy & 0.3043 & 0.3720 & 0.3558 & 0.4332 & 0.4497 & 0.4807 & 0.5201 & \textbf{0.5662} \\
      & Macro-F1 & 0.0985 & 0.1970 & 0.1863 & 0.2730 & 0.2853 & 0.3059 & 0.3164 & \textbf{0.3433} \\
      \hline
      \multirow{2}{*}{oag-venue} & Accuracy & 0.1361 & 0.1577 & 0.1392 & 0.2230 & 0.8359 & 0.8126 & 0.9615 & \textbf{0.9623} \\
      & Macro-F1 & 0.0681 & 0.1088 & 0.0878 & 0.1425 & 0.7628 & 0.7453 & 0.9057 & \textbf{0.9186} \\
      \hline
      \multirow{2}{*}{oag-field} & Accuracy & 0.4474 & 0.5017 & 0.5646 & 0.6757 & 0.6789 & 0.6804 & 0.6960 & \textbf{0.7178} \\
      & Macro-F1 & 0.1535 & 0.1939 & 0.2400 & 0.2831 & 0.3920 & 0.3887 & 0.3960 & \textbf{0.4074} \\
      \hline
    \end{tabular}
\end{table*}

\textbf{ogbn-mag}\footnote{\url{https://ogb.stanford.edu/docs/nodeprop/\#ogbn-mag}} The ogbn-mag dataset is a heterogeneous academic network constructed by Microsoft academic data, including four types of nodes: author (A), paper (P), field (F) and institution (I). The target nodes are papers, which are divided into 349 classes by their venue. Each paper node is associated with a 128-dimensional word2vec feature vector. For other types of nodes, we use node embeddings from metapath2vec \cite{metapath2vec} as its features. We use the original split from dataset.

\textbf{oag-venue} \cite{hgt} The oag-venue dataset is a heterogeneous academic network in Computer Science (CS) domain, which has the same graph schema with ogbn-mag. The labels of papers are their venues with 360 classes. The input features of papers and fields are 128-dimensional vectors calculated by the fine-tuned SciBERT \cite{beltagy2019scibert} model, while other types of nodes are represented by metapath2vec \cite{metapath2vec} vectors. Papers are split according to their published year.

\textbf{oag-field} We construct another heterogeneous academic network based on Microsoft academic data from Open Academic Graph\footnote{\url{https://www.aminer.cn/oag-2-1}}, called oag-field. The graph schema is the same as ogbn-mag. The labels of papers are their $L$1-level fields with 41 classes. The input feature of nodes and the split strategy are the same with oag-venue.


\subsection{Baselines}

We compare RHCO with several state-of-the-art heterogeneous graph neural network models:
\begin{itemize}
\item GAT \cite{velivckovic2018graph}, which employs the attention mechanism to select more important neighbors adaptively.
\item R-GCN\cite{rgcn}, which investigates  the relations in knowledge graphs by employing specialized transformation matrices for each type of edge.
\item C\&S \cite{cands}, which combines the advantages of label propagation and deep graph neural network to improve the node classification performance.
\item Mp2vec \cite{metapath2vec}, which uses the metapath-guided random walk to generate node sequences and perform a heterogeneous skip-gram algorithm to obtain node representations.
\item HGT \cite{hgt}, which utilizes type-specific parameters to capture the characteristics of different nodes and relations inspired by the Transformer~\cite{transformer}.
\item HGConv \cite{yu2020hybrid}, which performs convolutions at both micro and macro levels to learn the importance of nodes and relations respectively.
\item R-HGNN \cite{yu2022heterogeneous}, which investigates the role of relations for improving the learning of more fine-grained node representations.
\end{itemize}

The metapath-based methods HAN \cite{han} and HeCo \cite{heco} are not evaluated due to the difficulty in how to use sampling strategy to train these models with the constraints of multiple metapaths\cite{yu2022heterogeneous}.

\subsection{Experimental Setup}

We use PyTorch and DGL framework to implement RHCO and all baseline models.
Following \cite{han}, we test the performance of GAT on the graph generated by pre-defined metapaths (i.e., PAP, PFP and PPP on ogbn-mag and oag-venue, PAP, PVP and PPP on oag-field) and report the best performance.
For RHCO, we use HGT \cite{hgt} as the pre-trained model, hidden dimension of node embeddings $d = 64$, hidden dimension of relation embeddings $d_{rel} = 8$, number of attention heads $K = 8$, number of layers of network schema encoder $L = 2$, number of positive samples for each node $T_{pos} = 5$, dropout probability $p_{d}$ =  0.5, temperature parameter $\tau = 0.8$, contrastive loss balance coefficient $\lambda = 0.5$, contrastive loss weight $\alpha = 0.9$, number of label smoothing iterations $T = 50$.
For other baselines, we use the parameters in their original paper.

We use mini-batch training and neighbor sampling, batch size is 512, the number of sampled neighbors is 10, training for 150 epochs. We use Adam optimizer with 0.001 learning rate.

\begin{figure*}
    \begin{subfigure}{.3\textwidth}
      \centering
      \captionsetup{justification=centering}
      \includegraphics[width=1\linewidth]{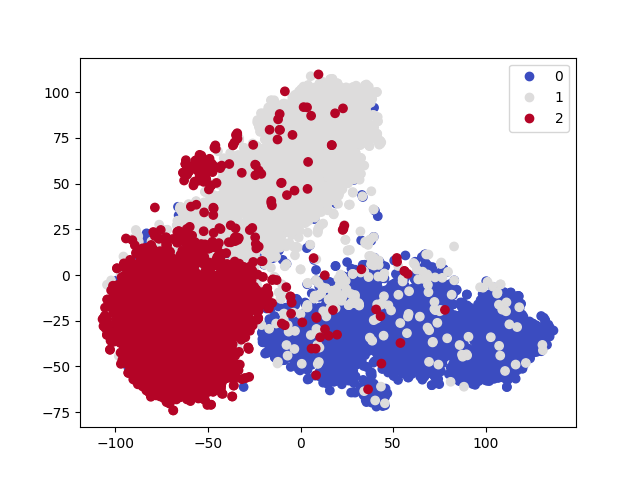}
      \vspace{-0.5cm}
      \caption{R-GCN}
    \end{subfigure}
    \begin{subfigure}{.3\textwidth}
      \centering
      \captionsetup{justification=centering}
      \includegraphics[width=1\linewidth]{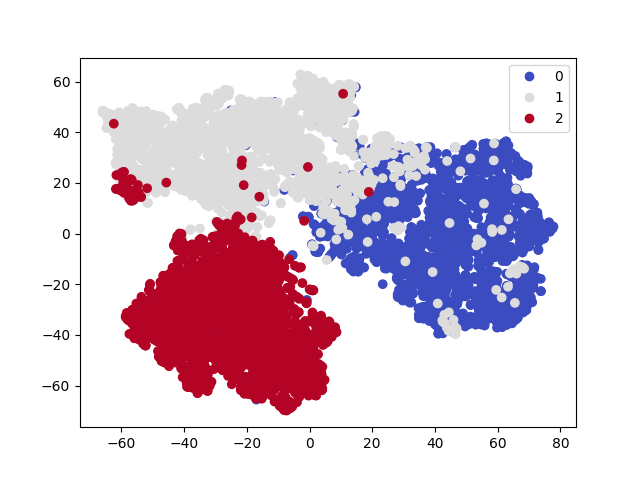}
      \vspace{-0.5cm}
      \caption{C\&S}
    \end{subfigure}
    \begin{subfigure}{.3\textwidth}
      \centering
      \captionsetup{justification=centering}
      \includegraphics[width=1\linewidth]{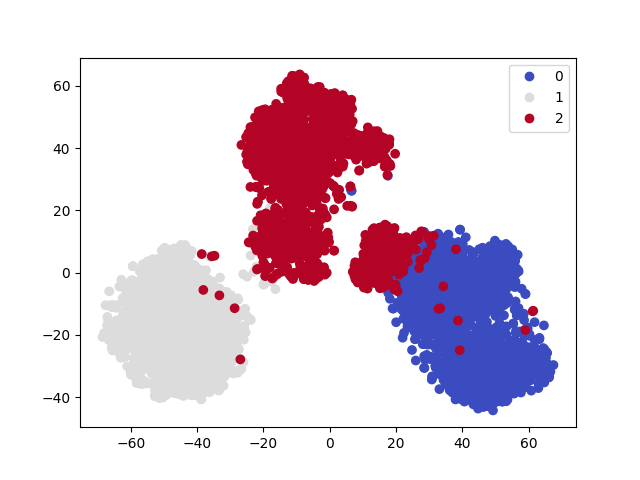}
      \vspace{-0.5cm}
      \caption{HGConv}
    \end{subfigure}
    \begin{subfigure}{.3\textwidth}
      \centering
      \captionsetup{justification=centering}
      \includegraphics[width=1\linewidth]{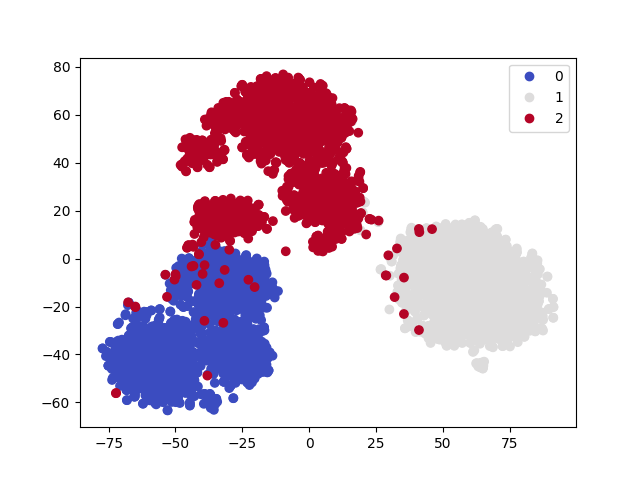}
      \vspace{-0.5cm}
      \caption{HGT}
    \end{subfigure}
    \begin{subfigure}{.3\textwidth}
      \centering
      \captionsetup{justification=centering}
      \includegraphics[width=1\linewidth]{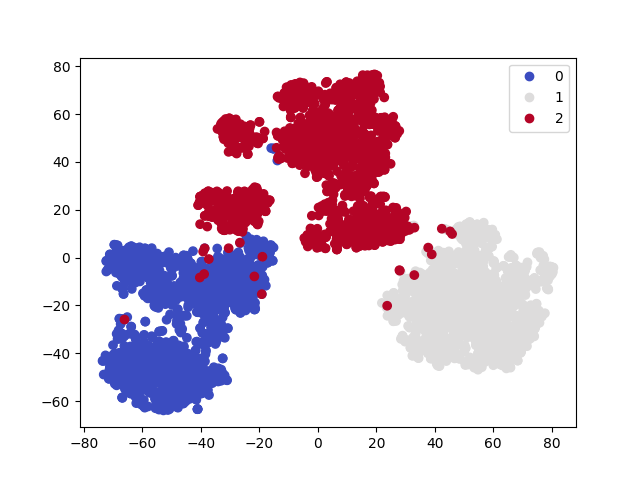}
      \vspace{-0.5cm}
      \caption{R-HGNN}
    \end{subfigure}
    \begin{subfigure}{.3\textwidth}
      \centering
      \captionsetup{justification=centering}
      \includegraphics[width=1\linewidth]{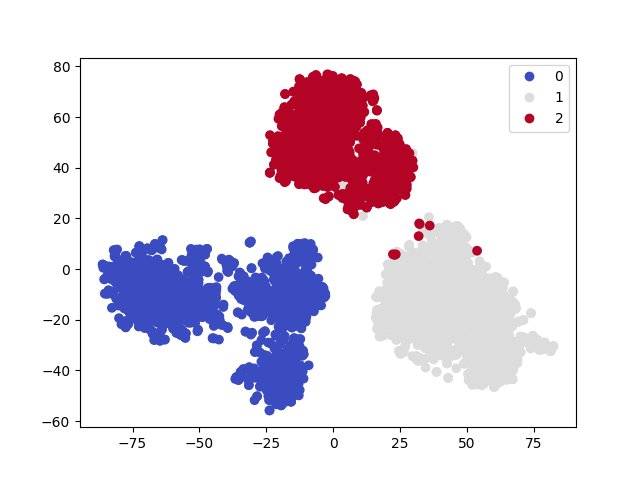}
      \vspace{-0.5cm}
      \caption{RHCO}
    \end{subfigure}
    \vspace{-0.3cm}
    \caption{Visualization of the node representations on ogbn-mag.}
\label{fig:visualization}
\end{figure*}

\begin{figure*}
    \begin{subfigure}{.3\textwidth}
      \centering
      \captionsetup{justification=centering}
      \includegraphics[width=1\linewidth]{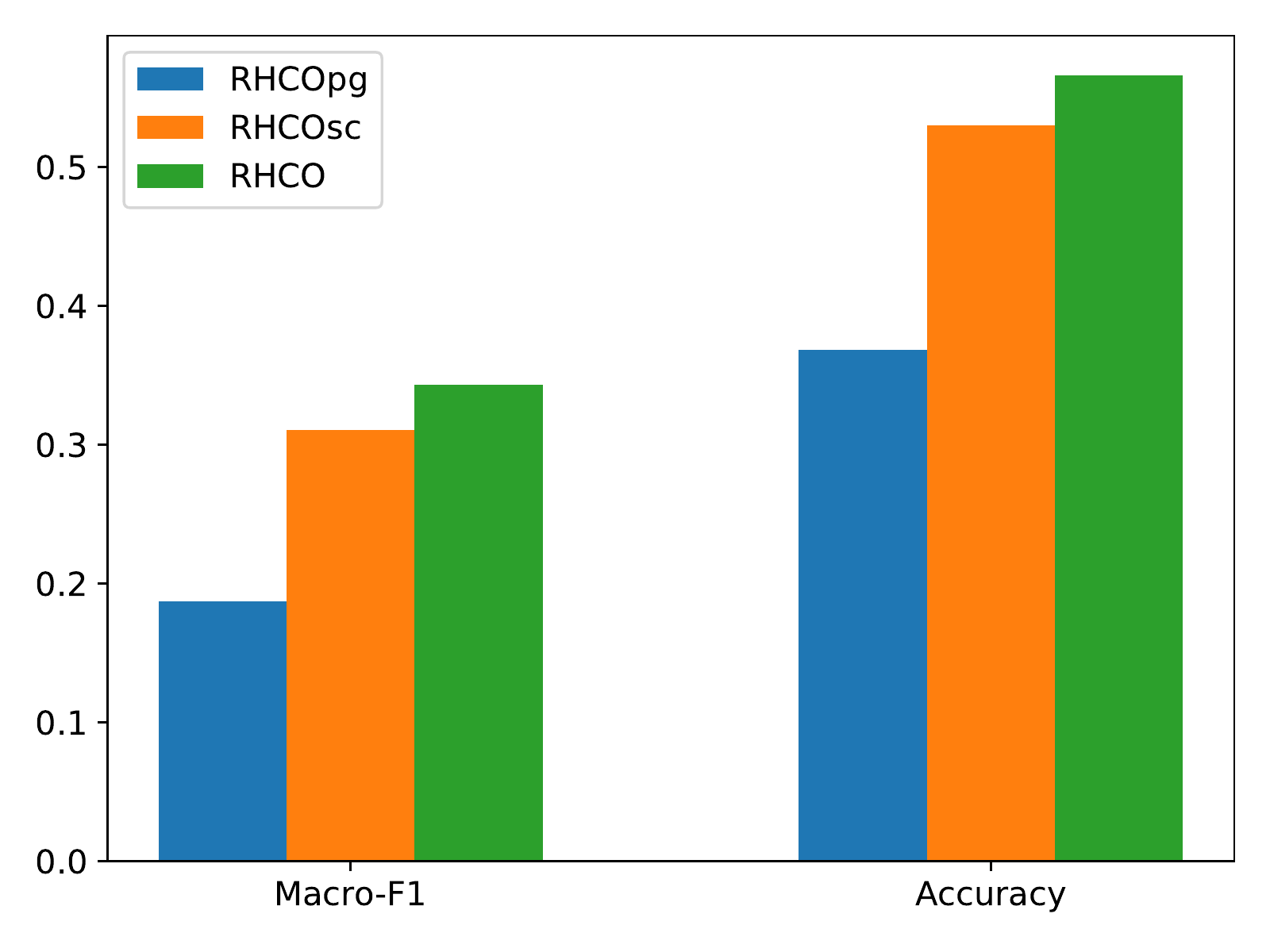}
      \vspace{-0.5cm}
      \caption{ogbn-mag}
    \end{subfigure}
    \begin{subfigure}{.3\textwidth}
      \centering
      \captionsetup{justification=centering}
      \includegraphics[width=1\linewidth]{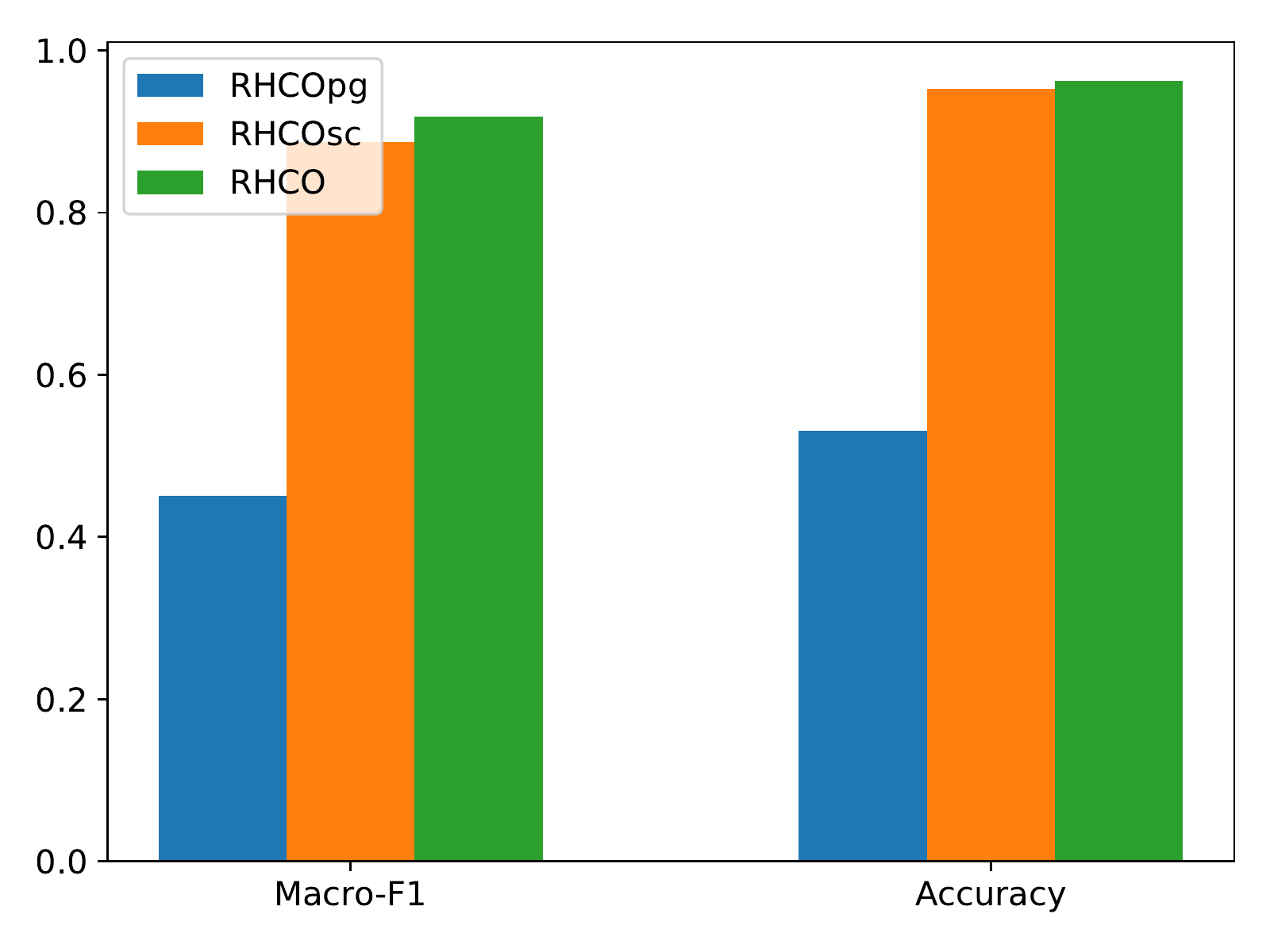}
      \vspace{-0.5cm}
      \caption{oag-venue}
    \end{subfigure}
    \begin{subfigure}{.3\textwidth}
      \centering
      \captionsetup{justification=centering}
      \includegraphics[width=1\linewidth]{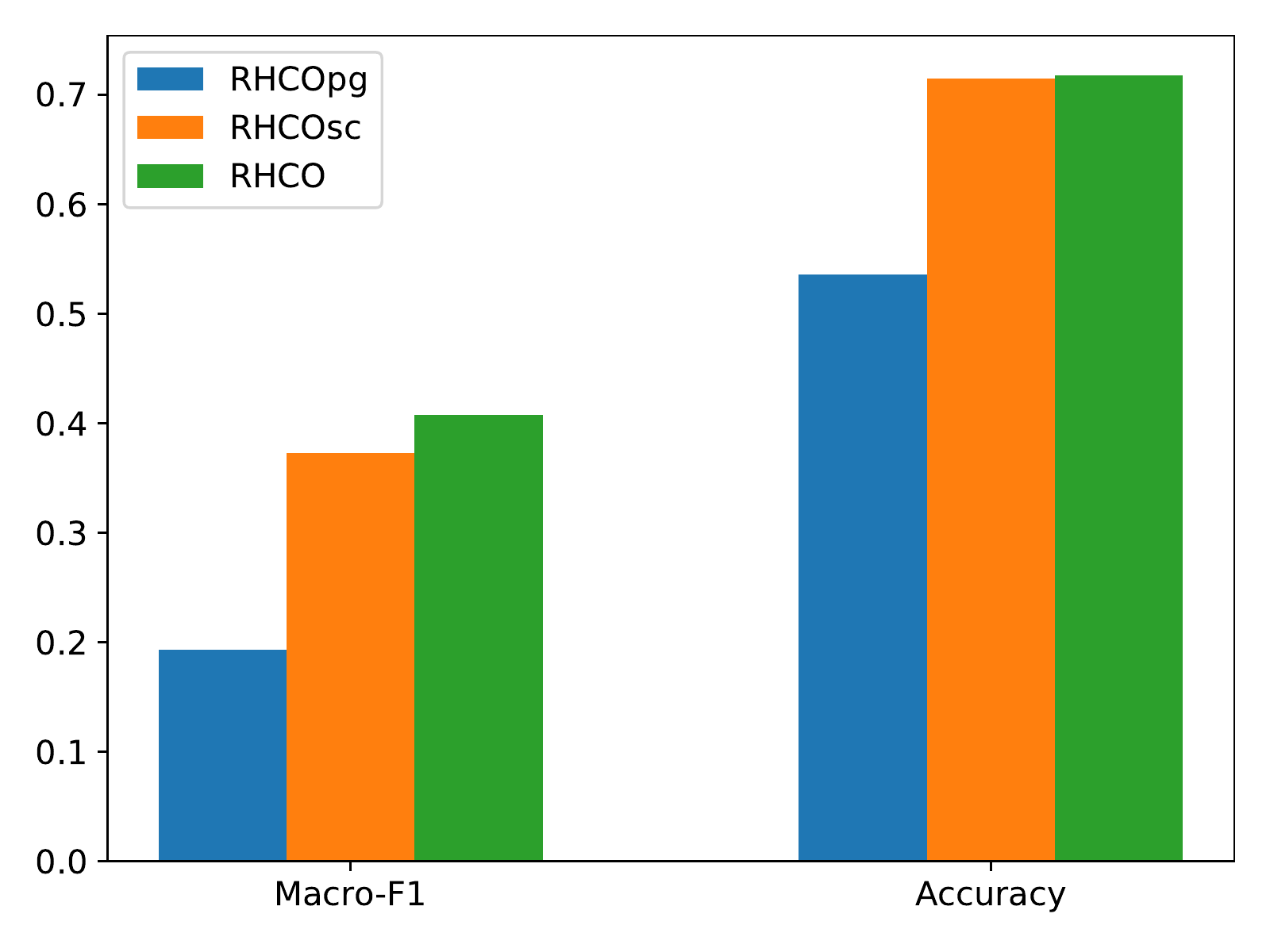}
      \vspace{-0.5cm}
      \caption{oag-field}
    \end{subfigure}
    \vspace{-0.3cm}
    \caption{Effects of network schema encoder and positive sample graph encoder.}
\label{fig:ablation_study}
\vspace{-0.5cm}
\end{figure*}

\begin{figure*}
   \begin{subfigure}{.3\textwidth}
     \centering
     \captionsetup{justification=centering}
     \includegraphics[width=1\linewidth]{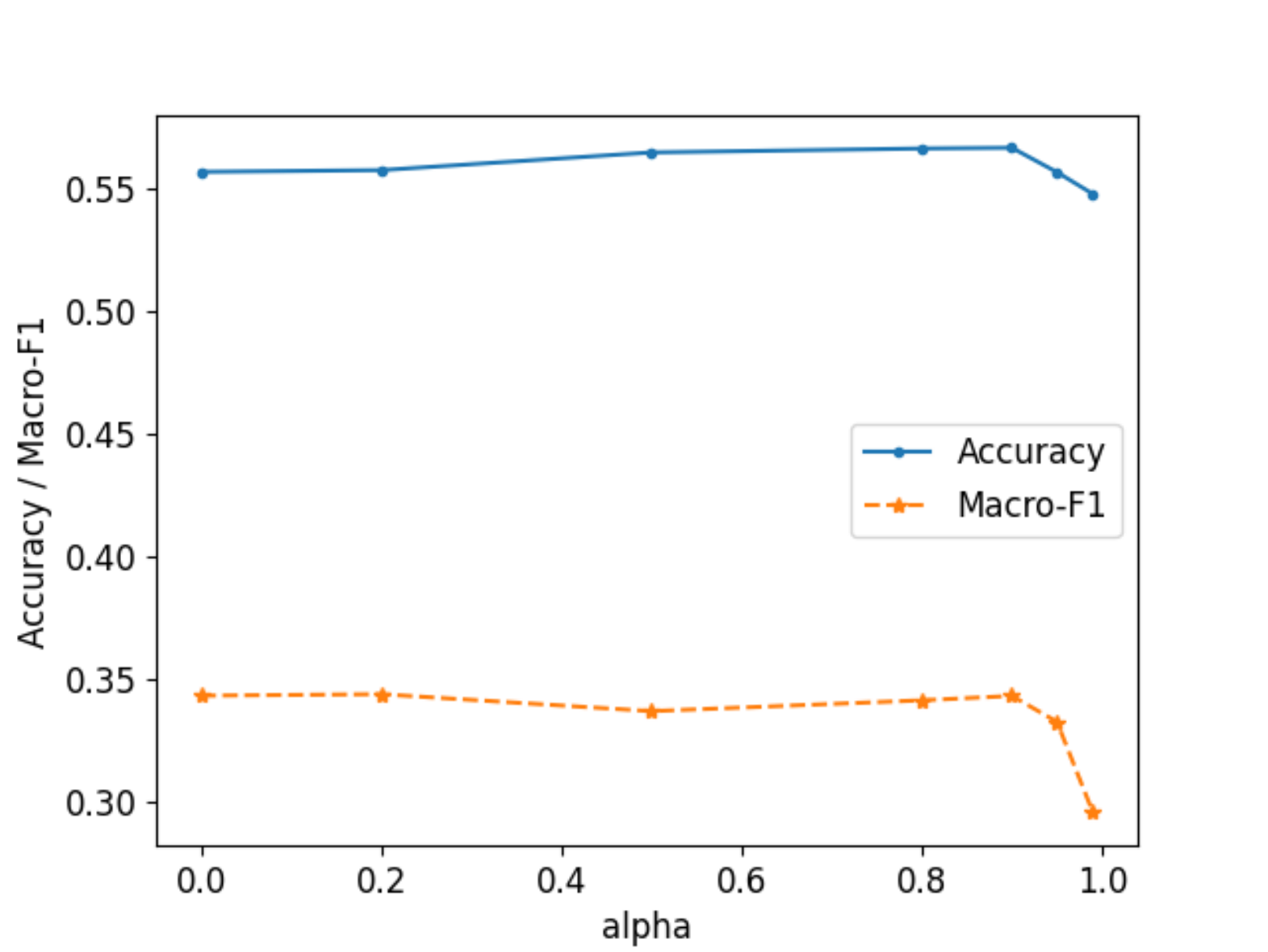}
     \caption{Contrastive Loss Weight}
   \end{subfigure}
   \begin{subfigure}{.3\textwidth}
     \centering
     \captionsetup{justification=centering}
     \includegraphics[width=1\linewidth]{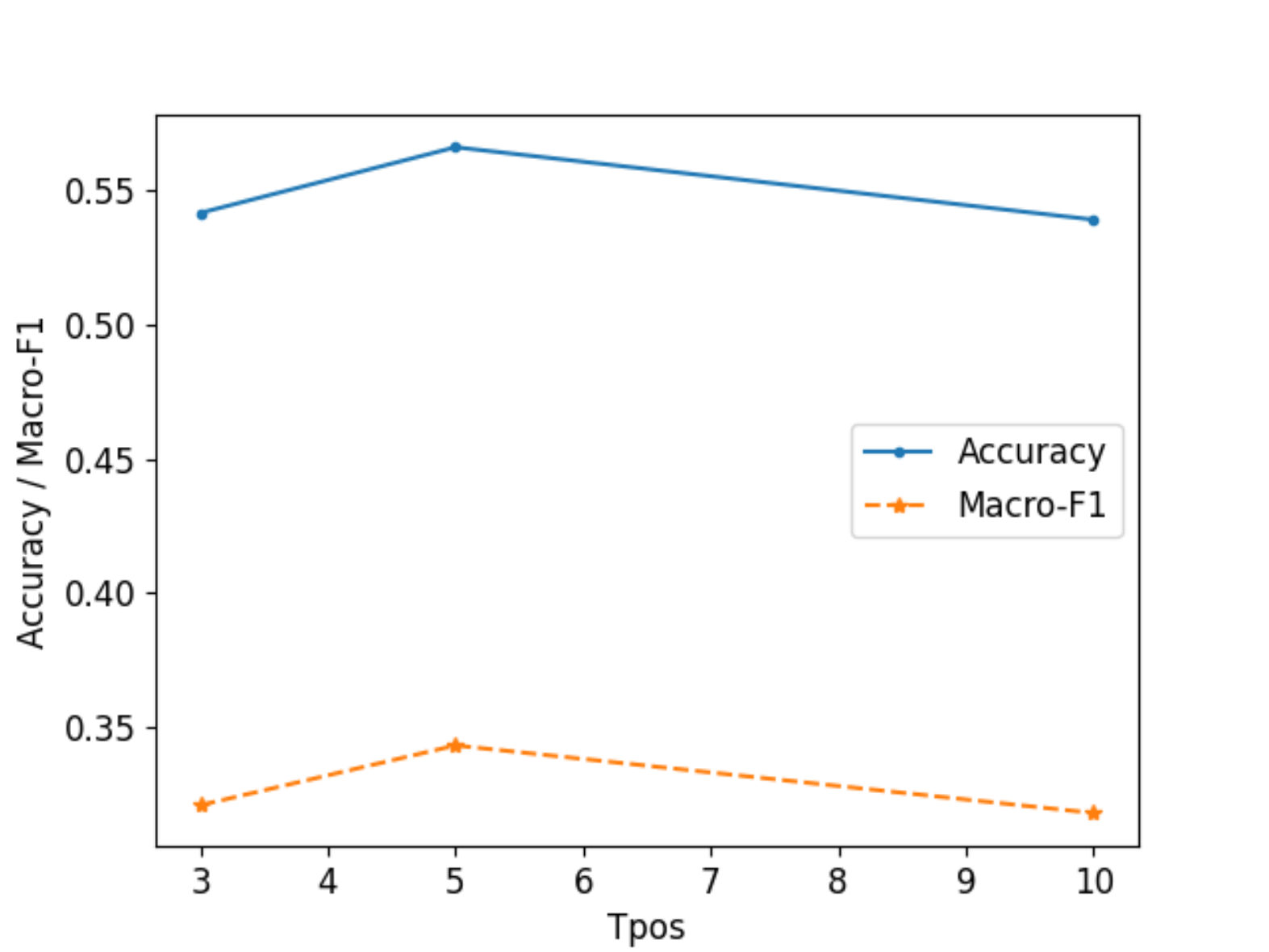}
     \caption{Number of Positive Samples}
   \end{subfigure}
   \begin{subfigure}{.3\textwidth}
     \centering
     \captionsetup{justification=centering}
     \includegraphics[width=1\linewidth]{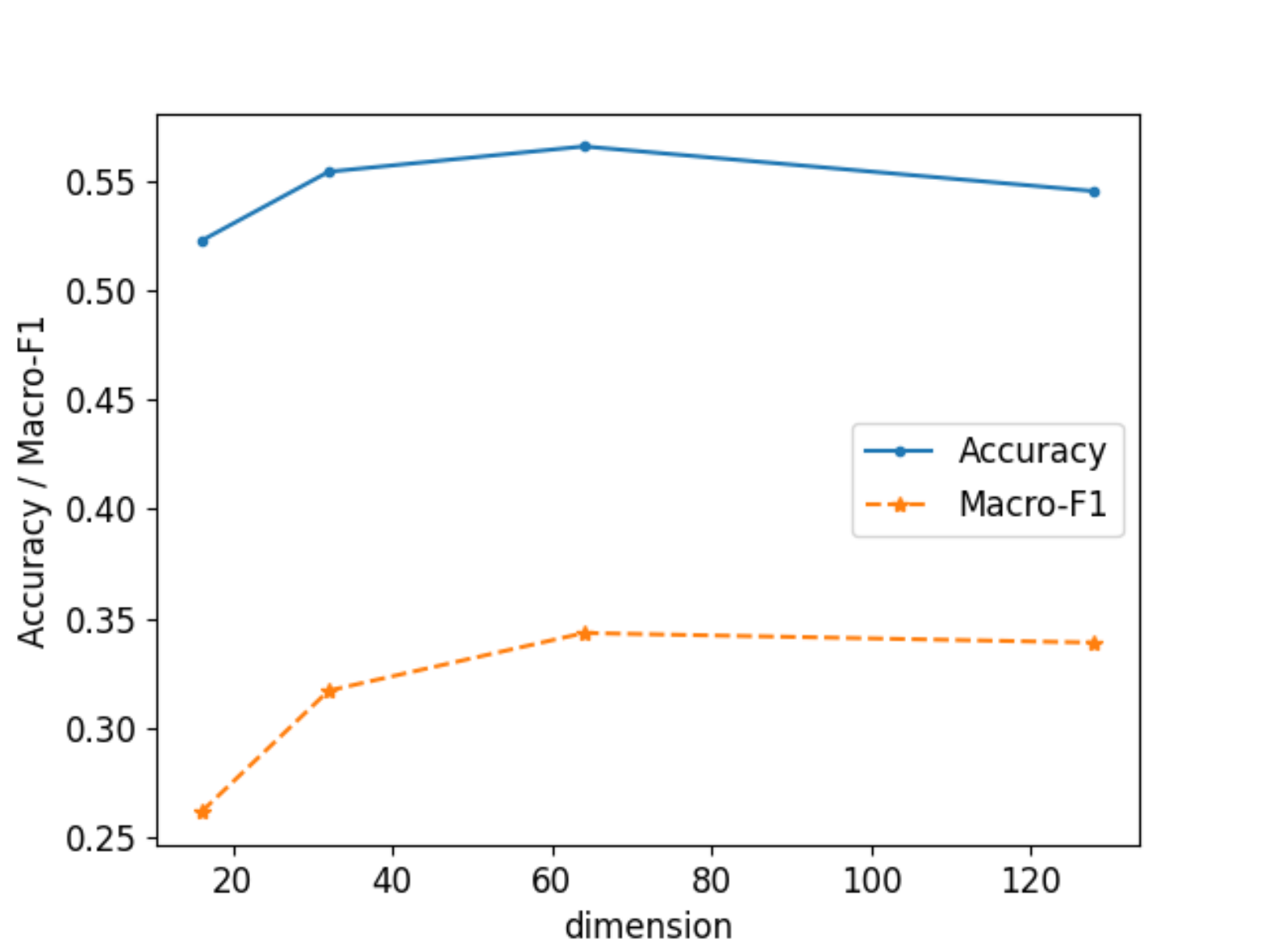}
     \caption{Dimension of Node Embeddings}
   \end{subfigure}
 \caption{Parameter Sensitivity Analysis on ogbn-mag}
 \label{fig:param_analysis}
\vspace{-0.3cm}
\end{figure*}

\subsection{Node Classification Result}
 
We use accuracy and macro-F1 to evaluate the performance of the above models on the node classification task. The results are shown in Table \ref{tab:result}.

As can be seen, RHCO outperforms all baselines on all datasets. By analyzing the results, several conclusions could be summarized.
Firstly, GAT performs worse than other methods, indicating that homogeneous graph neural networks failed to leverage the graph heterogeneity.
Secondly, R-GCN and C\&S both have bad performance on three datasets, which implies the necessity of using high-order information of large-scale graphs.
Moreover, HGT and HGConv perform worse than RHCO and R-HGNN, which shows that considering the role of relations in heterogeneous graph learning can capture the heterogeneity better and improve the learnt node embeddings.
Finally, compared with R-HGNN, our method achieves up to 4.6\% and 2.6\% improvement respectively. This shows that contrastive learning across views is more effective than single-view learning because cross-view contrastive learning could collaboratively supervise different views and learns more comprehensive node representations. 

In general, the observations demonstrate that RHCO can learn better node embeddings on large-scale, complex-structured graphs with the help of contrastive learning and the proposed positive sample selection strategy, which enables RHCO to construct positive samples on large-scale graphs.

\subsection{Node Visualization}

We conduct the visualization of node representations on ogbn-mag dataset to provide a more intuitive evaluation. Since there are 349 classes in ogbn-mag dataset, which cannot all be drawn in the results, we select node representations of top-three classes and project them into a 2-dimensional space using t-SNE\cite{2008Visualizing}. We plot learnt embeddings of R-GCN, C\&S, HGConv, HGT, R-HGNN and RHCO, and the results are shown in Figure \ref{fig:visualization}, where different colors denote different classes.

From Figure \ref{fig:visualization}, we find that R-GCN and C\&S present blurred boundaries between different classes of nodes because they failed to fuse all kinds of semantics. HGConv, HGT and R-HGNN both have some wrong classified nodes near the class center and nodes with the same category are not gathered together. Compared with baselines, RHCO has the most apparent boundaries and nodes with the same class are gathered closely.

\subsection{Ablation Study}

We design two variants RHCO$_{sc}$ and RHCO$_{pg}$ to verify the effectiveness of network schema encoder and positive sample graph encoder.
RHCO$_{sc}$ only uses network schema encoder to learn node embeddings, and the embeddings of positive and negative samples in contrastive loss also come from network schema encoder;
RHCO$_{pg}$ only uses positive sample graph encoder to learn node embeddings, and the embeddings of positive and negative samples in contrastive loss, as well as the embeddings for downstream tasks also come from positive sample graph encoder.
We compare these two variants and the overall RHCO model on the node classification task on all datasets and report the performance in Figure \ref{fig:ablation_study}.

From Figure \ref{fig:ablation_study}, we can conclude that the performance of the overall RHCO model is consistently better than the two variants, which proves the effectiveness and necessity of cross-view contrastive learning. 
The performance of RHCO$_{sc}$ is also very competitive and has only a small gap with RHCO in oag-venue and oag-field dataset, which shows that aggregating information from different types of first-order neighbors can use multiple types of proximity and obtain better node embeddings than from high-order neighbors of the same type.
The bad performance of RHCO$_{pg}$ indicates that the metapath-based method is not sufficient to capture the heterogeneity of graphs in large-scale datasets even though it can be applied to large-scale datasets using the proposed positive sample selection strategy. Therefore, although metapath is a widely used and powerful tool in heterogeneous graph representation learning, its application scope is limited to small-scale datasets.

\subsection{Parameter Sensitivity Analysis}

We analyze the influence of the three parameters of RHCO on ogbn-mag dataset: contrastive loss weight $\alpha$, number of positive samples $T_{pos}$, and dimension of node embeddings $d$. We conduct experiments on the node classification task in ogbn-mag dataset. The results are shown in Figure \ref{fig:param_analysis}.

\textbf{Contrastive loss weight}. We vary the contrastive loss weight $\alpha$ from 0 to 1 and report the result in Figure \ref{fig:param_analysis} (a). The result shows that the model performance is better at $0.2 \leq \alpha \leq 0.9$ than $\alpha = 0$ (without contrastive learning), and the best performance is achieved at $\alpha = 0.9$. This shows that combining semi-supervised learning with contrastive learning helps the model learn better node embeddings; when $0.9 < \alpha <1$, the performance drops significantly, which shows that it is difficult to learn effective embeddings for downstream tasks by relying too much on unsupervised contrastive learning.

\textbf{Number of positive samples}. We change the number of positive samples $T_{pos}$ and report the result in Figure \ref{fig:param_analysis} (b). $T_{pos}$ determines the number of neighbors in Eq. \ref{eq:pg_encoder_gcn} and the size of the positive sample set in Eq. \ref{eq:contrastive_loss}. As $T_{pos}$ increases, the performance of the model will first increase, because target nodes can gather more information from positive samples in positive sample graph encoder. However, when $T_{pos}$ is too large, the label consistency between the target node and its positive samples will decrease, which will affect the performance of label propagation and result in a decrease in the final performance.

\textbf{Dimension of node embeddings}. We explore the effect of the dimension of node embeddings $d$ and show the result in Figure \ref{fig:param_analysis} (c). It can be seen that as $d$ increases, the performance of the model first increases and then decreases, reaching the highest at $d = 64$. This shows that RHCO needs a suitable embedding dimension to encode information. If $d$ is too small, the expression ability is insufficient, and when $d$ is too large, additional redundant information will be introduced.

\section{Conclusion}

In this paper, we studied the problem of heterogeneous graph representation learning in large-scale graphs and proposed a heterogeneous graph neural network model based on cross-view contrastive learning called RHCO.
RHCO first applies contrastive learning to large-scale heterogeneous graphs. RHCO learns node embeddings from network schema encoder and positive sample graph encoder respectively and utilizes cross-view contrastive learning to supervise the learnt embeddings.
For the scalability problem of existing methods, RHCO adopts a novel positive sample selection strategy, which uses pre-trained attention values instead of the number of connected metapaths to construct positive sample pairs. 
In the network schema encoder, RHCO also considers the role of various relations in heterogeneous graphs and learns relation-aware node representations, which improves the learnt node representations. 
Extensive experimental results demonstrate the performance improvement of RHCO in heterogeneous graph representation learning.


\begin{acks}
To Robert, for the bagels and explaining CMYK and color spaces.
\end{acks}

\balance
\bibliographystyle{ACM-Reference-Format}
\bibliography{sample-base}

\end{document}